\documentclass{article} 
\usepackage{nips14submit_e,times}
\usepackage{hyperref}
\usepackage{latexsym}
\usepackage{url}
\usepackage{booktabs}
\usepackage{graphicx}
\usepackage[ruled,vlined]{algorithm2e}
\usepackage{subfig}
\usepackage{caption}
\usepackage{amsmath,amssymb,calc,ifthen,capt-of}
\usepackage{multirow}

\newcommand{\nobracket}{}

\newcommand{\tmmathbf}[1]{\ensuremath{\boldsymbol{#1}}}

\newcommand{\tmop}[1]{\ensuremath{\operatorname{#1}}}

\title{Bayesian Optimisation for Machine Translation}

\author{
Yishu Miao$^1$   Ziyu Wang$^1$    Phil Blunsom$^{1,2}$ \\
$^1$Department of Computer Science, University of Oxford\\
$^2$DeepMind Technologies\\
\texttt{\{yishu.miao,ziyu.wang,phil.blunsom\}@cs.ox.ac.uk} \\
}

\nipsfinalcopy 

\begin{document}
\maketitle
\begin{abstract}
  This paper presents novel Bayesian optimisation algorithms for minimum error rate training of statistical machine translation systems. We explore two classes of algorithms for efficiently exploring the translation space, with the first based on N-best lists and the second based on a hypergraph representation that compactly represents an exponential number of translation options. Our algorithms exhibit faster convergence and are capable of obtaining lower error rates than the existing translation model specific approaches, all within a generic Bayesian optimisation framework. Further more, we also introduce a random embedding algorithm to scale our approach to sparse high dimensional feature sets. 

\end{abstract}

\section{Introduction}

State of the art statistical machine translation (SMT) models traditionally consist of a small number ($<$20) of sub-models whose scores are linearly combined to choose the best translation candidate.
The weights of this linear combination are usually trained to maximise some automatic translation metric (e.g. BLEU) \cite{Papineni2002} using Minimum Error Rate Training (MERT) \cite{Och2002,och2003minimum} or a variant of the Margin Infused Relaxed Algorithm (MIRA) \cite{Crammer2003,Chiang2012}.
These algorithms are heavily adapted to exploit the properties of the translation search space.
In this paper we introduce generic, effective, and efficient Bayesian optimisation (BO) algorithms \cite{brochu2010tutorial,Snoek2012} for training the weights of SMT systems for arbitrary metrics that outperform both MERT and MIRA.
To our knowledge this is the first application of BO in natural language processing (NLP) and our results show that their may be significant scope for using BO to tune hyperparameter in a range of NLP models.

The linear model popular for SMT systems \cite{Och2002} is parametrised in terms of a source sentence $\tmmathbf{f}$, target translation $\tmmathbf{e}$, feature weights $w_k$ and corresponding feature functions $H_k ( \tmmathbf{e}, \tmmathbf{f})$ (including a language model, conditional translation probabilities, etc.). The best translation is selected by,
\begin{eqnarray}
\hat{\tmmathbf{e}} =  \arg \underset{\tmmathbf{e}}{\max} \{ \sum_{k = 1}^K w_k H_k ( \tmmathbf{e}, \tmmathbf{f}) \}. 
\end{eqnarray}
Since the translation metrics (e.g. BLEU score) can only be evaluated between the selected translations and reference translations (i.e. the standard manual translations from the parallel training data), meanwhile decoding new translations following Equation 1 is very time consuming, we cannot tune the linear weights directly as in ordinary classification tasks.
The most common approach is an iterative algorithm MERT \cite{och2003minimum} which employs N-best lists (the best N translations decoded with a weight set from a previous iteration) as candidate translations $\tmmathbf{C}$. In this way, the loss function is constructed as $E ( \tmmathbf{\bar{E}}, \hat{\tmmathbf{E}}) = \sum_{s = 1}^S E ( \tmmathbf{\bar{e}}_s, \hat{\tmmathbf{e}}_s)$, where $\tmmathbf{\bar{e}}$ is the reference sentence, $\hat{\tmmathbf{e}}$ is selected from N-best lists by $\hat{\tmmathbf{e}}_s = \arg \underset{\tmmathbf{e} \in \tmmathbf{C}}{\max} \left\{ \sum_{k = 1}^K w_k H_k ( \tmmathbf{e}, \tmmathbf{f}_s) \right\}$ and $S$ represents the volume of sentences. By exploiting the fact that the error surface is piece-wise linear, MERT iteratively applies line search to find the optimal parameters along the randomly chosen directions via Equation 2, generating new N-best lists until convergence (no change happened in the new N-best lists),
\begin{eqnarray}
  \hat{\tmmathbf{w}} & = & \arg \underset{\tmmathbf{w}}{\min}
  \left\{ \sum_{s = 1}^S E \left( \tmmathbf{\bar{e}}_s, \arg \underset{\tmmathbf{e}
  \in \tmmathbf{C}}{\max} \left\{ \sum_{k = 1}^K w_k H_k ( \tmmathbf{e},
  \tmmathbf{f}_s) \right\} \right) \right\} .
\end{eqnarray}

Hypergraph, or lattice, MERT{ \cite{Macherey2008,Kumar2009} aims to tackle
problems caused by the lack of diversity in N-best lists.
A hypergraph \cite{huang2008advanced} efficiently encodes the exponential translation space explored by the beam-search translation decoder.
The line search can then be carried out on the edges of the hypergraph, instead of the translations in the N-best lists. And dynamic programming is used to find the upper envelope of the hypergraph corresponding to the maximum scoring translation. Prior work \cite{Macherey2008,Kumar2009} showed that hypergraph MERT is superior to the original N-best algorithm both in speed of convergence and stability. 
MIRA is an online large-margin learning algorithm that applies a different strategy to MERT. It enforces a margin between high and low loss translations and enables stochastic gradient descent to be used to update parameters. A disadvantage of this approach is that it requires the global BLEU score, which is a non-linear function of local translation candidate statistics, to be approximated by a linear combination of sentence level BLEU scores.

In this paper, however, our BO algorithms treat the loss function as a black-box function so that we could directly query the function value without the cumbersome and inefficient work of constructing an error surface for random directions. Instead of applying BO to the whole SMT pipeline, which would require expensive decoding of new translations with every parameter set sampled, our BO algorithms only decode new translations after obtaining the best parameters on fixed N-best lists or hypergraphs. Hence our algorithms iteratively run Gaussian processes on the sub-models and only a few decoding iterations are required to reach convergence. The experiments in Section 3 illustrate the superiority of our algorithms both in translation quality and speed of convergence.

\begin{figure}[!b]
\begin{minipage}[b]{0.5\textwidth}
\raisebox{3cm}{
\begin{algorithm}[H]
 \footnotesize
 \SetKwInOut{Input}{Input}
 \SetKwInOut{Output}{Output}
 \Input{Initial weights $\tmmathbf{w}_0$, source sentences $\tmmathbf{F}$, reference sentences $\tmmathbf{\bar{E}}$.}
 \Output{Final weights $\tmmathbf{w}_{f}$}
 \For{$i=0;i<maxIter;i=i+1$}{
  Decode hypergraphs $\tmmathbf{H}_i$ using $\tmmathbf{w}_i$;\\
  Generate search bound $\tmmathbf{B}_i$= $\{ w_i^k \in \tmmathbf{w}_i | \nobracket w_i^k - b \leqslant x_i^k \leqslant w_i^k + b \}$;\\
  Initialise candidate point set $\mathcal{X}$ in bounded search area $\tmmathbf{B}_i$ of the Gaussian process;\\
  \For{$j=0;j<maxBOIter;j=j+1$}{
  	$\tmmathbf{x}_j = \arg \underset{\tmmathbf{x} \in \mathcal{X}}{\max} \tmmathbf{EI} ( \tmmathbf{x} )$;\\
    Reweight hypergraphs $\tmmathbf{H}_i$ by $\tmmathbf{x}_j$;\\
  	Generate translation set $\tmmathbf{\hat{E}}_j$ by Viterbi algorithm;\\
  	$y_j = \tmop{BLEU} (\tmmathbf{\bar{E}},\tmmathbf{\hat{E}}_j)$;\\
    Update GP with $(\tmmathbf{x}_j,y_j)$;\\
  } 
  $\tmmathbf{w}_{i + 1} =\tmmathbf{x}_{\tmop{\emph{best}}}$;\\  
 }
 Return $\tmmathbf{w}_i$
 \caption{Hypergraph BO}
 \label{alg:hg}
\end{algorithm}
}
\end{minipage}
\hspace{0.16cm}
\begin{minipage}[b]{0.6\textwidth}
  \includegraphics[bb=20 20 520 390, width=0.8\linewidth]{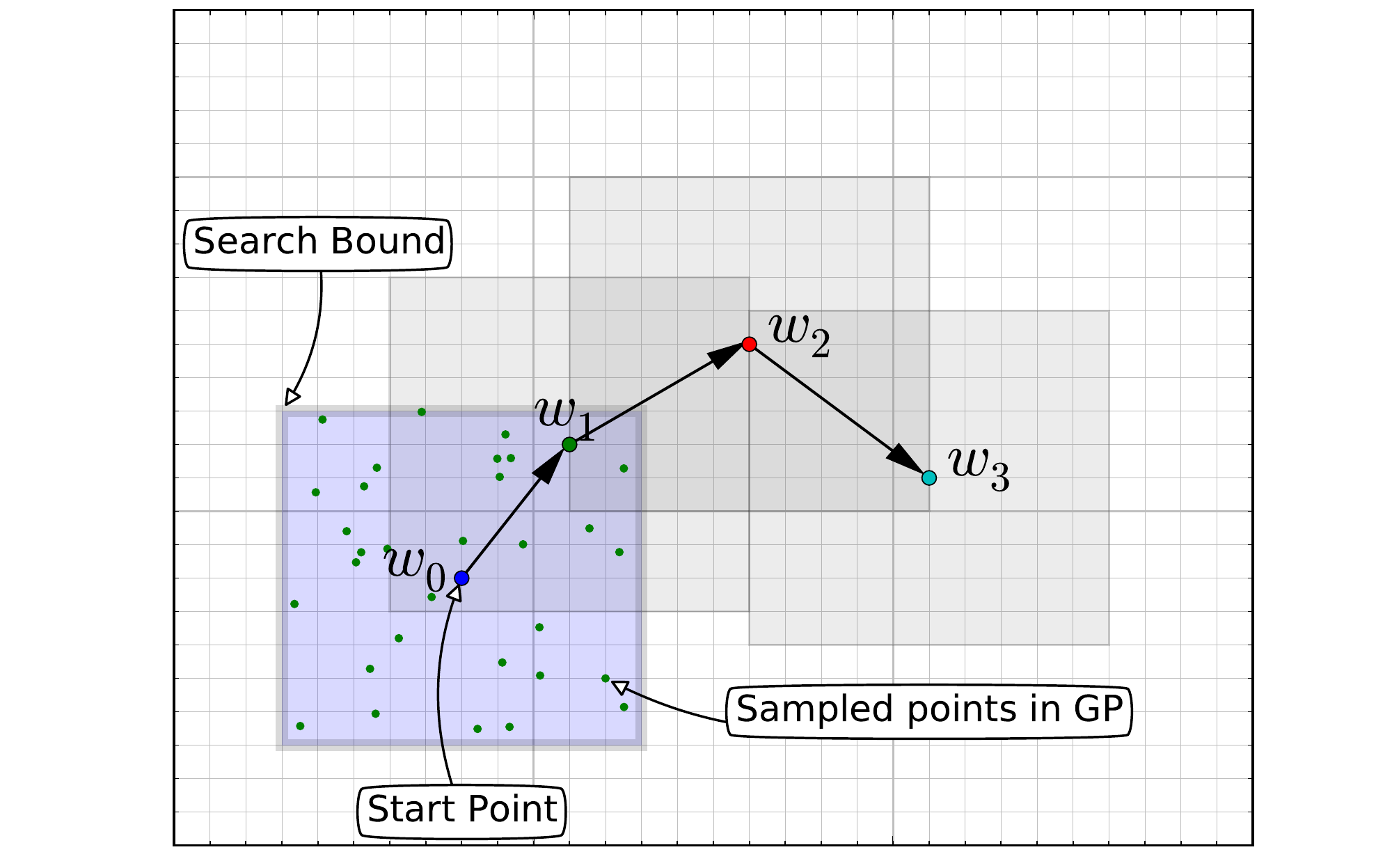}
\caption{Bounded search in 2 dimensions. \quad\quad\quad\quad\quad}
\label{fig:bound}
\end{minipage}
\end{figure}

\section{Bayesian Optimisation Tuning Algorithms}
Algorithm \ref{alg:hg} describes our hypergraph algorithm (HG-BO). 
The N-best algorithm (NBL-BO) is similar to HG-BO and can be derived from Algorithm \ref{alg:hg} by replacing the hypergraphs with N-best lists. 
In HG-BO, both $\tmmathbf{w}_i$ and $\tmmathbf{x}_j$ represent the weights of the linear model. 
The weights $\tmmathbf{w}_i$ are used to produce the hypergraphs $\tmmathbf{H}_i$, while $\tmmathbf{x}_j$ are the weights sampled from the GP to compute the BLEU score (i.e. objective function value) for a fixed set $\tmmathbf{H}_i$. Since $\tmmathbf{H}_i$ remains unchanged during an iteration of Bayesian optimisation, the BLEU score calculated for the fixed hypergraphs approximates the true BLEU score that would be achieved if the translation system were run with $\tmmathbf{x}_j$. This introduces some noise owing to the variance between $\tmmathbf{w}_i$ and $\tmmathbf{x}_j$.

As depicted in Fig. \ref{fig:bound}, a key aspect of Algorithm \ref{alg:hg} is that we place a bound (blue area) around $\tmmathbf{w}_i$ and only consider samples inside this region. 
The sample with the highest BLEU score will then be used to decode new hypergraphs for the next iteration of BO. 
Intuitively, to speed up convergence, we would like the search space of BO to be as large as possible. 
When the search space is too large, however, a sampled $\tmmathbf{x}_j$ could be so far from $\tmmathbf{w}_i$
that the generated translations would become unreliable thus leading to noisy BLEU measurements. 
HG-BO is preferable to NBL-BO as it weighs the translations directly in the hypergraphs, which encode an exponentially larger space of translations than the N-best lists, and thus noise is diminished. 
To further expand the translation space searched at each iteration, we present a variant cumulative hypergraph BO algorithm (CHG-BO) which combines hypergraphs from one previous and current iterations in order to trade stability and speed of convergence with memory usage.

Similar to MERT, our BO algorithms become less reliable when the number of features in the linear model exceeds 30. Hence, we introduce a variant of random embedding Bayesian optimisation (REMBO) \cite{Wang2013} into our hypergraph algorithm (HG-REMBO) to tackle the large scale training problem. The original REMBO generates a random matrix $\tmmathbf{A} \in {R}^{h \times l}$ to map the sample $\tmmathbf{x} \in {R}^h$ from high dimensional space to a point $\tmmathbf{z} \in {R}^l$ in low dimensional space. The objective function to be optimised then becomes $g ( \tmmathbf{z}) = f ( \tmmathbf{A}\tmmathbf{z})$. Instead of $\mathbf{A}$, we used a regularised random matrix
$\tmmathbf{\bar{A}}$ where $\bar{A}_{m n} = \frac{A_{m n}}{\| \tmmathbf{A}_m \|_1}$ and transform the objective function to $  g ( \tmmathbf{z}) = f ( \tmmathbf{\bar{A}} \tmmathbf{z}+\tmmathbf{w})$, where $\tmmathbf{w}$ are the weights producing the hypergraphs. $\tmmathbf{w}$ would remain constant during Bayesian optimisation. 
In this way, BO can be carried out in the low dimensional space
and the regularisation of $\mathbf{A}$ ensures that each update of the weights remains in a bounded domain.

\section{Experiments}
We implemented our models using \emph{spearmint} \cite{Snoek2012}\footnote{https://github.com/JasperSnoek/spearmint} and the \emph{cdec} SMT decoder{ \cite{Dyer2010}}\footnote{http://www.cdec-decoder.org/}. 
The datasets are from WMT14 shared task,\footnote{http://www.statmt.org/wmt14/translation-task.html} 
all tokenized and lowercased. We employ ARD Matern 5/2 kernel and  EI acquisition function. 
The \emph{cdec} implementations of hypergraph MERT \cite{Kumar2009} and MIRA\cite{Chiang2009} are used as benchmarks.    

\begin{table}[t]
\setlength{\abovecaptionskip}{0.1cm}
\setlength{\belowcaptionskip}{-0.3cm} 
\centering
\scriptsize
\addtolength{\tabcolsep}{-4.5pt}
\begin{tabular}{c|clcc|clcc|clcc|clcc}
  \toprule[1.2pt]
  Language& \multicolumn{4}{c|}{French-English (fr-en)} &  \multicolumn{4}{c|}{Spanish-English (es-en)} & \multicolumn{4}{c|}{German-English (de-en)}&\multicolumn{4}{c}{Czech-English (cs-en)}\\
  [1pt]
  \hline
  Dataset & Dev&(variance) & Test-1 &Test-2 & Dev&(variance) & Test-1& Test-2 & Dev&(variance) & Test-1 &Test-2 & Dev&(variance) & Test-1 & Test-2\\
  [1pt]
  \hline
  MERT 
  & 26.1 &\tiny(2$\times$10$^{-2}$)& 26.8 & 26.5
  & 29.5 &\tiny(\textbf{1}$\times$\textbf{10}$^{-4}$)& 28.2 & 30.2 
  & 21.1 &\tiny(6$\times$10$^{-1}$)& 18.9 & 20.2
  & 17.7 &\tiny(5$\times$10$^{-1}$)& 17.8 & 16.9\\
  MIRA 
  & 26.0 &\tiny(1$\times$10$^{-3}$)& 26.8 & 26.5
  & 29.2 &\tiny(1$\times$10$^{-3}$)& \textbf{28.5} & \textbf{30.7}
  & 20.9 &\tiny(1$\times$10$^{-2}$)& 18.9 & 20.3
  & 18.4 &\tiny(4$\times$10$^{-3}$)& 18.7 & 17.7 \\
  [1pt]
  \hline
  NBL-BO 
  & \textbf{26.4} &\tiny(6$\times$10$^{-5}$)& 26.7 & 26.5
  & 29.7 &\tiny(1$\times$10$^{-2}$)& 28.1 & 30.4
  & 22.0 &\tiny(2$\times$10$^{-3}$)& \textbf{19.8} & \textbf{21.0}
  & 18.8 &\tiny(\textbf{2}$\times$\textbf{10}$^{-3}$)& 18.8 & 17.3\\
  HG-BO 
  & \textbf{26.4} &\tiny(\textbf{3}$\times$\textbf{10}$^{-5}$)& 26.8 & 26.7
  & \textbf{29.9} &\tiny(\textbf{1}$\times$\textbf{10}$^{-4}$)& 28.0 & 30.1
  & \textbf{22.2} &\tiny(\textbf{1}$\times$\textbf{10}$^{-5}$)& \textbf{19.8} & 20.9
  & 19.1 &\tiny(1$\times$10$^{-2}$)& 19.1 & 17.7\\
  CHG-BO 
  & \textbf{26.4} &\tiny(3$\times$10$^{-3}$)& \textbf{26.9} & \textbf{26.8}
  & \textbf{29.9} &\tiny(2$\times$10$^{-3}$)& 28.3 & 30.4  
  & 22.1 &\tiny(3$\times$10$^{-2}$)& 19.7 & 20.9 
  & \textbf{19.2} &\tiny(\textbf{2}$\times$\textbf{10}$^{-3}$)& \textbf{19.3} & \textbf{17.8}\\
  \bottomrule[1.2pt]
\end{tabular}
\caption{Translation Performance (BLEU) score}
\label{tbl:bleu}
\end{table}

\begin{figure}[t]
\setlength{\abovecaptionskip}{0.1cm}
\setlength{\belowcaptionskip}{-0.5cm} 
\raisebox{0.1cm}{
\begin{minipage}[t]{0.65\textwidth} 
\centering
\begin{tabular}{lc}
  \subfloat[\emph{fr-en}]{\label{fig:2-a}\includegraphics[bb=16 0 520 390,width=0.48\linewidth]{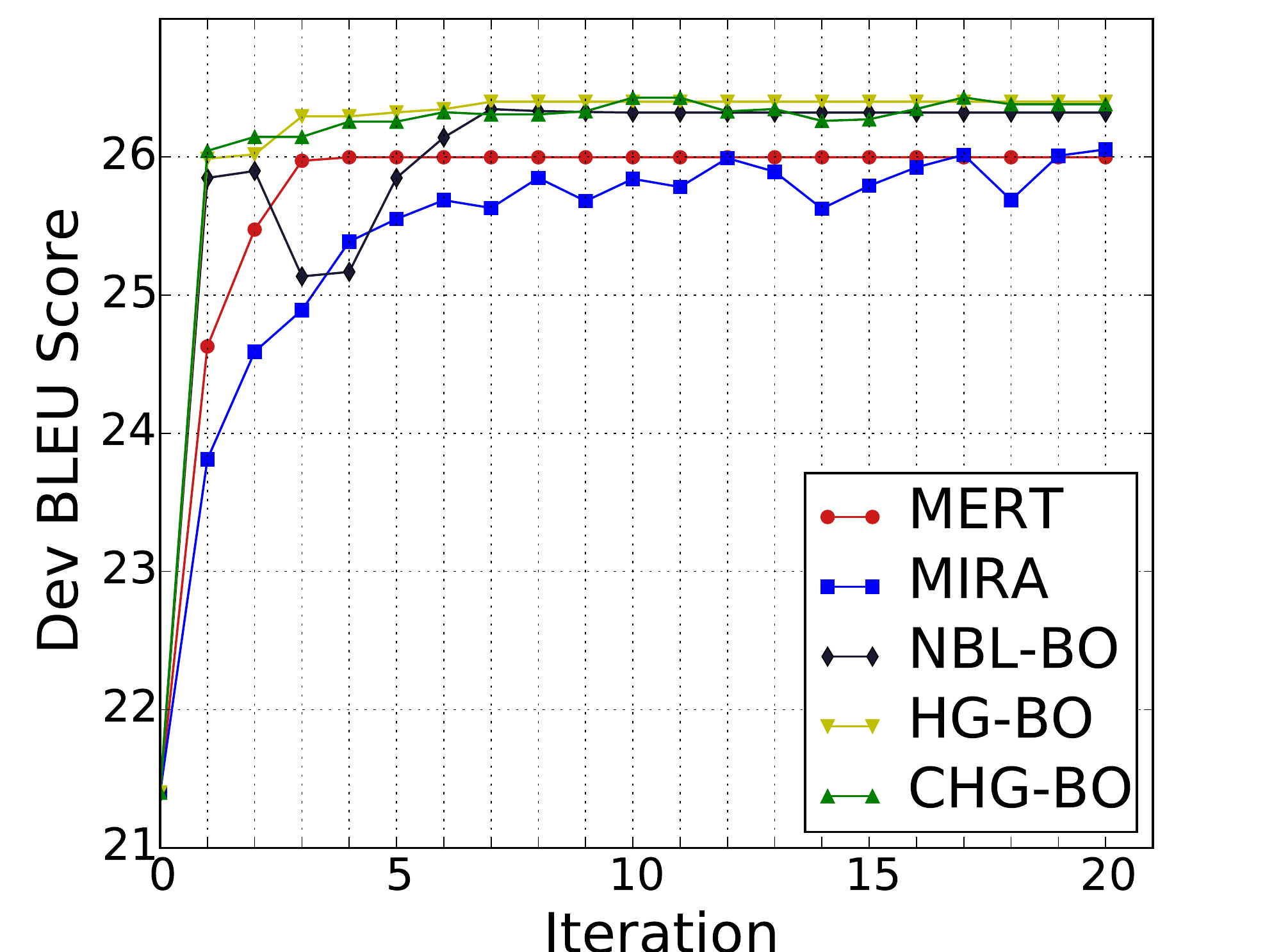}}
  \hspace{0.2cm}
  \subfloat[\emph{cs-en}]{\label{fig:2-b}\includegraphics[bb=16 0 520 395,width=0.48\linewidth]{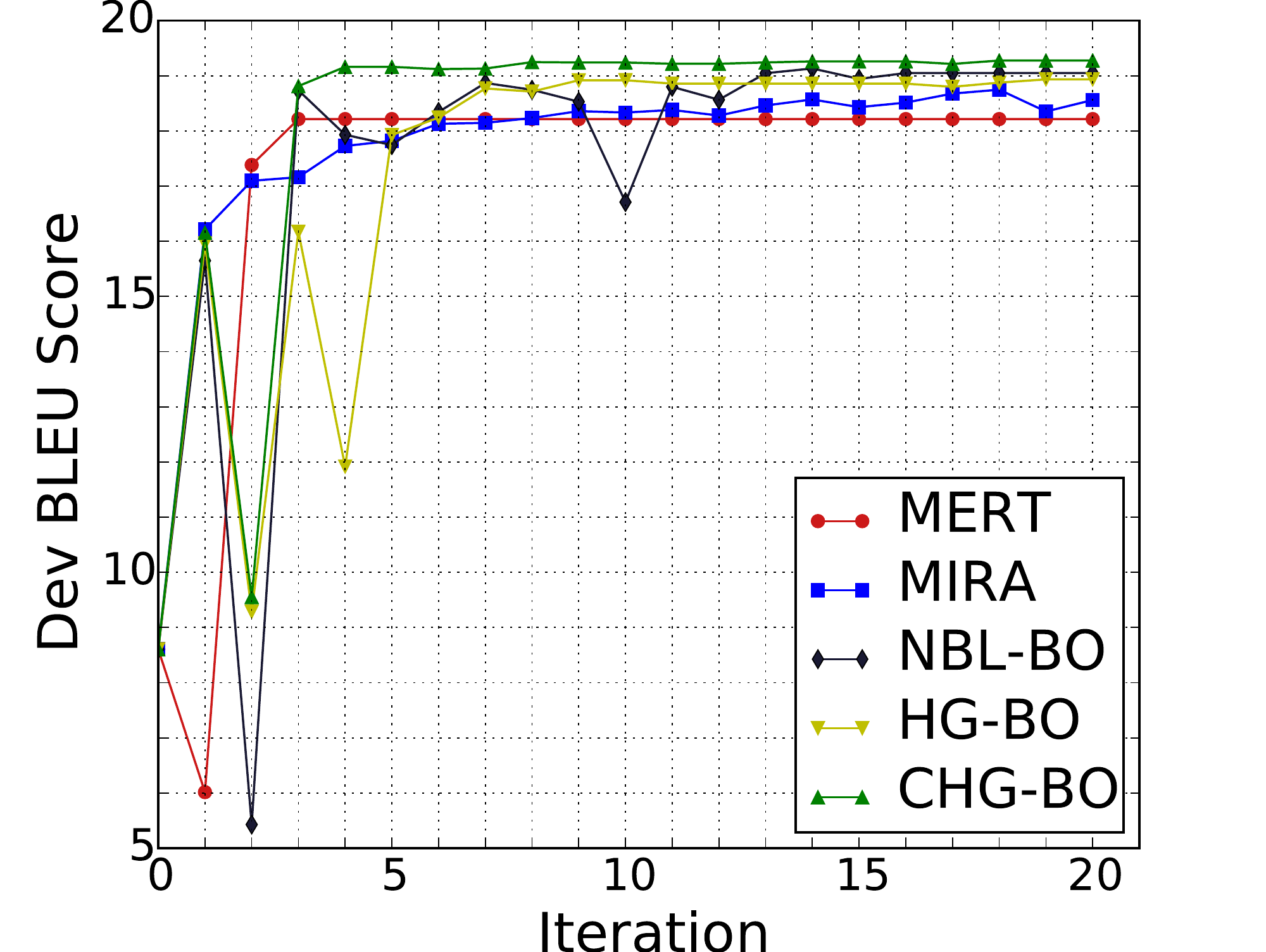}} 
 \end{tabular}
\caption{Convergence of different models}
\label{fig:dev}
\end{minipage} 
}
\begin{minipage}[t]{0.35\textwidth} 
\addtolength{\tabcolsep}{-5.0pt}
  \centering
  \footnotesize
  \begin{tabular}{c|c|c}
  \toprule[1.2pt]
    Model& Time(h) & Iteration\\
    \hline
    MERT & 4 & 5\\
    MIRA & 4 & 20\\
    \hline
    NBL-BO & 1.5 & 5\\
    HG-BO & 1.5 & 5\\
    CHG-BO & 2 & 5\\
    \bottomrule[1.2pt]
  \end{tabular}
  \captionof{table}{Time consumption}
  \label{tbl:time}
 \end{minipage} 
\end{figure}

The experiment\footnote{The 4-gram language model is trained on \emph{europarl}, \emph{news-crawl} and \emph{news-commentary} sections, translation grammar is extracted from \emph{news-commentary}, while \emph{news-test} 2010 is used for BO, \emph{news-test} 2011 and 2012 are used for testing. We use 18 default \emph{cdec} features and the same initial weights on one machine with 10 processors and trained for 20 iterations. 
The BO bound size is 0.1 and the number of BO iterations is 100.} 
results in Table \ref{tbl:bleu}, averaged over 3 runs, show that our BO algorithms always achieve a higher training objective score than MERT and MIRA, and in most cases a higher test BLEU score. 
Fig.\ref{fig:dev} illustrates the convergence w.r.t.\ the development BLEU score and Fig. \ref{fig:2-b} shows a particular case where the imperfect starting weights cause a violent fluctuation initially. 
CHG-BO quickly reaches the plateau in 5 iterations but NBL-BO dips again at the 10th iteration. 

Table \ref{tbl:time} illustrates the efficiency of the BO algorithms. 
They consistently obtain a good weight set within 5 iterations, but the best one is always achieved after 7 iterations. 
This suggests setting the maximum number of iterations to 10 in order to ensure a good result. 
Our BO tuning algorithms only take advantage of multiple processors for decoding, thus there still exists some space to further improve their efficiency.          

\begin{figure}[t]
\setlength{\abovecaptionskip}{-0.01cm} 
\setlength{\belowcaptionskip}{-0.3cm}
\captionsetup[subfigure]{labelformat=empty}
\addtolength{\tabcolsep}{-2.8pt}
	\begin{tabular}{ccc}
	  \subfloat[(a) Convergence of \emph{fr-en}]{\label{fig:3-a}\includegraphics[bb=26 13 520 390,width=0.30\linewidth]{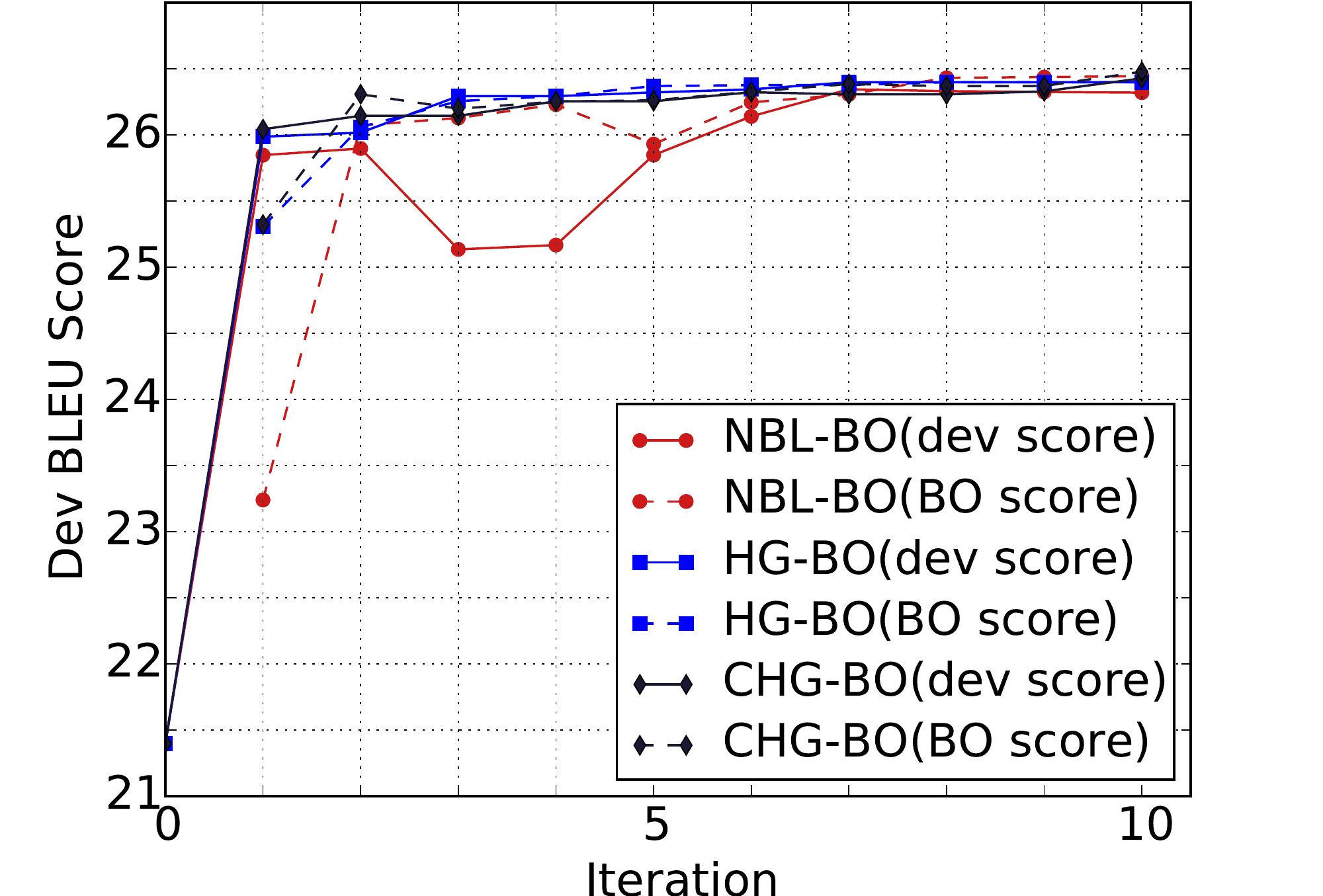}}
	  &
	  \subfloat[(b) Convergence of \emph{cs-en}]{\label{fig:3-b}\includegraphics[bb=26 13 520 390,width=0.30\linewidth]{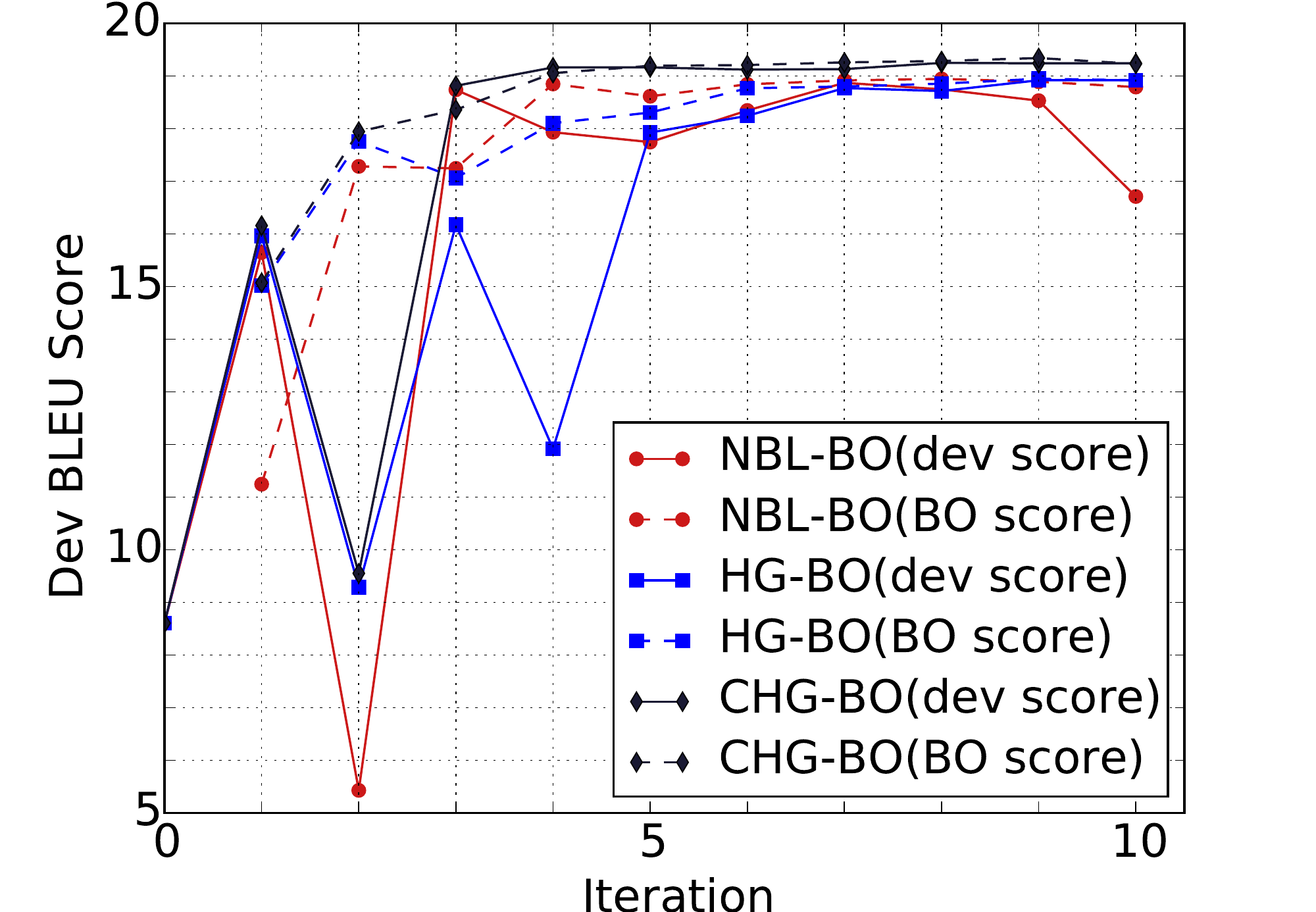}} 
	  &
	  \subfloat[(c) Different bound size of \emph{fr-en} ]{\label{fig:4-a}\includegraphics[bb=26 13 520 390,width=0.30\linewidth]{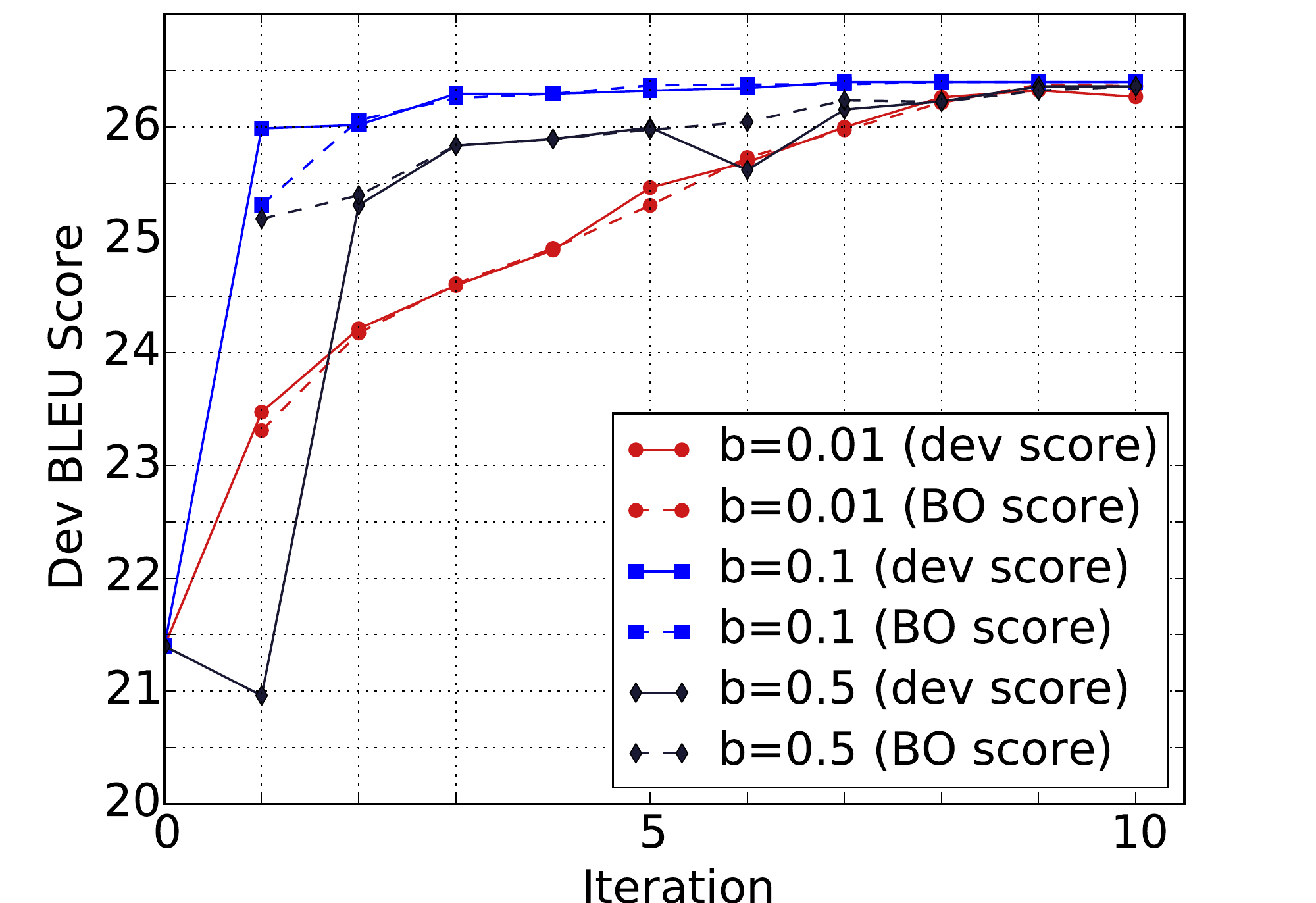}}
	  \\  
	  \subfloat[(d) Test BLEU score of \emph{fr-en}]{\label{fig:4-b}\includegraphics[bb=26 13 520 390,width=0.30\linewidth]{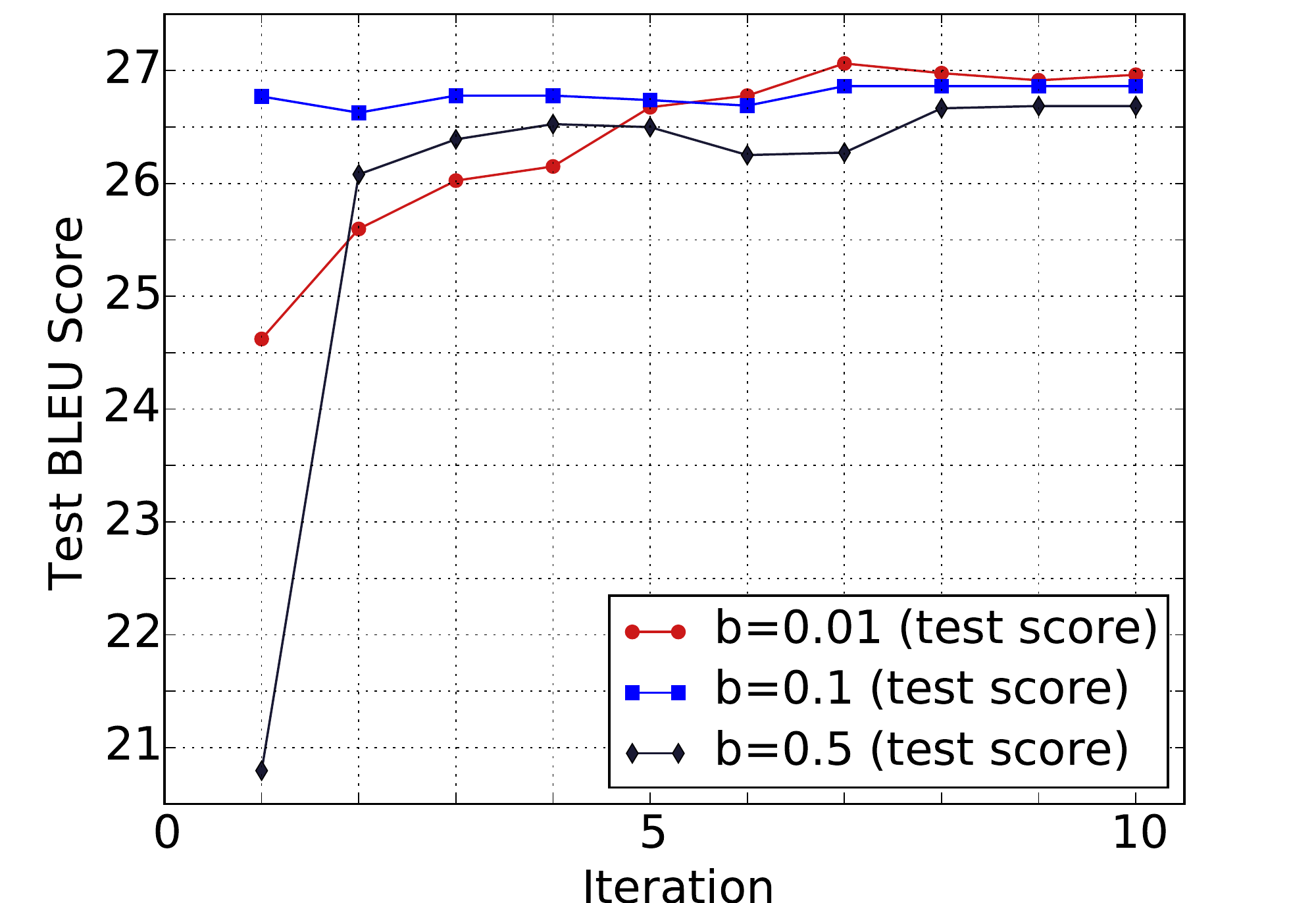}} 
	   &
	 \multicolumn{2}{c}{
	  \raisebox{2.3cm}{\subfloat[Table 3: Adding four sparse feature functions, 1) rule identifier, 2) rule shape, 3) bigrams on source side and 4) target side of rules, increases the number of features from 18 to 56,396 in \emph{es-en} and 207,952 in \emph{fr-en}. More details about the rules can be found in \cite{Simianer2012}.]{
		\footnotesize
		\addtolength{\tabcolsep}{-2.0pt}
		  \begin{tabular}{c|ccc|ccc}
		    \toprule[1.2pt]
		    Model & \multicolumn{3}{c|}{French (fr-en) } & \multicolumn{3}{c}{Spanish-English (es-en)} \\
		    \hline
		    Dataset &Dev (variance)&Test-1&Test-2&Dev (variance)&Test-1&Test-2 \\
		    \hline
			MIRA 
			&\textbf{26.8} \scriptsize(\textbf{3}$\times$\textbf{10}$^{-3}$)&\textbf{27.0}&26.5
			&\textbf{30.0} \scriptsize(4$\times$10$^{-3}$)&\textbf{28.3}&\textbf{30.6}	\\	    
		    HG-REMBO 
		    &26.5 \scriptsize(2$\times$10$^{-2}$)&26.9&\textbf{26.8}
		    &29.8 \scriptsize(\textbf{1}$\times$\textbf{10}$^{-5}$)&\textbf{28.3}&\textbf{30.6} \\
		    \bottomrule[1.2pt]
		  \end{tabular}
		  }
		}
	}
	\end{tabular}
\caption{Convergence of BO algorithms\quad\quad\quad\quad\quad\quad\quad\quad\quad\quad\quad\quad\quad\quad\quad\quad\quad\quad\quad\quad\quad\quad\quad\quad\quad\quad}
\label{tbl:rembo}
\end{figure}

Fig. \ref{fig:3-a} and \ref{fig:3-b} indicate the comparison of development score and BO score\footnote{BO score is the best BLEU score achieved by Gaussian processes on fixed N-best lists or hypergraphs.} at each iteration in \emph{fr-en} and \emph{cs-en}, which again demonstrates the advantage of CHG-BO on stability over NBL-BO and HG-BO. Fig. \ref{fig:4-a} and \ref{fig:4-b} compare the models with different bound size: $b=0.01$ is able to achieve a development and test BLEU score as good as $b=0.1$ with more iterations, but $b=0.5$ performs worse on the test dataset. 
Thus too large search bound may introduce too much noise which in turn affects the translation performance.   

Table \ref{tbl:rembo} shows the experiments on a large number of sparse features. We modify HG-REMBO into a two step coordinate ascent processes in order to stabilise the updates of the core default feature weights. First, we optimise the default 18 features, then we fix them and generate a regularised random matrix to update the large scale sparse features in the low dimensional space. Table \ref{tbl:rembo} demonstrates that HG-REMBO is able to carry out large scale discriminative training, performing almost on par with MIRA. Although HG-REMBO loses its advantage on speed of convergence as it requires multiple runs to generate a good transformation matrix, these results illustrate the potential of applying REMBO on statistical machine translation systems.
\vspace{-0.5em}
\section{Conclusion}
\vspace{-0.5em}
We introduce novel Bayesian optimisation (BO) algorithms for machine translation. 
Our algorithms exhibit faster convergence and achieve higher training objectives and better translation quality than existing translation model specific approaches. We further demonstrate that by incorporating the method of random embeddings it is viable to employ Bayesian optimisation to carry out large sale training with a high number of sparse features. This initial investigation also suggests that BO has great potential for general natural language processing tasks.
\vspace{-0.3em}
\section{Acknowledgements}
\vspace{-0.2em}
This work was supported by a Xerox Foundation Award and EPSRC grant number EP/K036580/1.

\newpage
\bibliographystyle{ieeetr}
\bibliography{TermPaper}

\begin{thebibliography}{10}

\bibitem{Papineni2002}
K.~Papineni, S.~Roukos, T.~Ward, and W.-J. Zhu, ``Bleu: a method for automatic
  evaluation of machine translation,'' in {\em Proceedings of the 40th Annual
  Meeting on Association for Computational Linguistics}, pp.~311--318,
  Association for Computational Linguistics, 2002.

\bibitem{Och2002}
F.~J. Och and H.~Ney, ``Discriminative training and maximum entropy models for
  statistical machine translation,'' in {\em Proceedings of the 40th Annual
  Meeting on Association for Computational Linguistics}, pp.~295--302,
  Association for Computational Linguistics, 2002.

\bibitem{och2003minimum}
F.~J. Och, ``Minimum error rate training in statistical machine translation,''
  in {\em Proceedings of the 41st Annual Meeting on Association for
  Computational Linguistics-Volume 1}, pp.~160--167, Association for
  Computational Linguistics, 2003.

\bibitem{Crammer2003}
K.~Crammer and Y.~Singer, ``Ultraconservative online algorithms for multiclass
  problems,'' {\em The Journal of Machine Learning Research}, vol.~3,
  pp.~951--991, 2003.

\bibitem{Chiang2012}
D.~Chiang, ``Hope and fear for discriminative training of statistical
  translation models,'' {\em The Journal of Machine Learning Research},
  vol.~13, no.~1, pp.~1159--1187, 2012.

\bibitem{brochu2010tutorial}
E.~Brochu, V.~M. Cora, and N.~De~Freitas, ``A tutorial on bayesian optimization
  of expensive cost functions, with application to active user modeling and
  hierarchical reinforcement learning,'' {\em arXiv preprint arXiv:1012.2599},
  2010.

\bibitem{Snoek2012}
J.~Snoek, H.~Larochelle, and R.~P. Adams, ``Practical bayesian optimization of
  machine learning algorithms,'' in {\em Advances in Neural Information
  Processing Systems}, pp.~2951--2959, 2012.

\bibitem{Macherey2008}
W.~Macherey, F.~J. Och, I.~Thayer, and J.~Uszkoreit, ``Lattice-based minimum
  error rate training for statistical machine translation,'' in {\em
  Proceedings of the Conference on Empirical Methods in Natural Language
  Processing}, pp.~725--734, Association for Computational Linguistics, 2008.

\bibitem{Kumar2009}
S.~Kumar, W.~Macherey, C.~Dyer, and F.~Och, ``Efficient minimum error rate
  training and minimum bayes-risk decoding for translation hypergraphs and
  lattices,'' in {\em Proceedings of the Joint Conference of the 47th Annual
  Meeting of the ACL and the 4th International Joint Conference on Natural
  Language Processing of the AFNLP: Volume 1-Volume 1}, pp.~163--171,
  Association for Computational Linguistics, 2009.

\bibitem{huang2008advanced}
L.~Huang, ``Advanced dynamic programming in semiring and hypergraph
  frameworks,'' {\em COLING, Manchester, UK}, 2008.

\bibitem{Wang2013}
Z.~Wang, M.~Zoghi, F.~Hutter, D.~Matheson, and N.~De~Freitas, ``Bayesian
  optimization in high dimensions via random embeddings,'' in {\em Proceedings
  of the Twenty-Third international joint conference on Artificial
  Intelligence}, pp.~1778--1784, AAAI Press, 2013.

\bibitem{Dyer2010}
C.~Dyer, J.~Weese, H.~Setiawan, A.~Lopez, F.~Ture, V.~Eidelman,
  J.~Ganitkevitch, P.~Blunsom, and P.~Resnik, ``cdec: A decoder, alignment, and
  learning framework for finite-state and context-free translation models,'' in
  {\em Proceedings of the ACL 2010 System Demonstrations}, pp.~7--12,
  Association for Computational Linguistics, 2010.

\bibitem{Chiang2009}
D.~Chiang, K.~Knight, and W.~Wang, ``11,001 new features for statistical
  machine translation,'' in {\em Proceedings of Human Language Technologies:
  The 2009 Annual Conference of the North American Chapter of the Association
  for Computational Linguistics}, pp.~218--226, Association for Computational
  Linguistics, 2009.

\bibitem{Simianer2012}
P.~Simianer, S.~Riezler, and C.~Dyer, ``Joint feature selection in distributed
  stochastic learning for large-scale discriminative training in smt,'' in {\em
  Proceedings of the 50th Annual Meeting of the Association for Computational
  Linguistics: Long Papers-Volume 1}, pp.~11--21, Association for Computational
  Linguistics, 2012.

\bibitem{krause2011contextual}
A.~Krause and C.~S. Ong, ``Contextual gaussian process bandit optimization,''
  in {\em Advances in Neural Information Processing Systems}, pp.~2447--2455,
  2011.

\end{thebibliography}

\end{document}